\documentclass[runningheads,a4paper]{llncs}

\usepackage{amssymb}
\usepackage{graphicx}
\usepackage{subfig}

\usepackage{url} 
\newcommand{\keywords}[1]{\par\addvspace\baselineskip
\noindent\keywordname\enspace\ignorespaces#1}

\begin{document}

\mainmatter  

\title{Development of Low-Cost Real-Time Driver Drowsiness Detection System using Eye Centre Tracking and Dynamic Thresholding}

\titlerunning{ }

%
%
\author{Fuzail Khan, Sandeep Sharma \and M.R Arulalan}

\authorrunning{ }

\institute{Department of Electronics and Communication Engineering, \\National Institute of Technology Karnataka, Surathkal, India - 575025} 

%
%

\toctitle{Development of Low-Cost Real-Time Driver Drowsiness Detection System using Eye Centre Tracking and Dynamic Thresholding}
\tocauthor{Fuzail Khan, Sandeep Sharma}

\maketitle

\begin{abstract}
One in every five vehicle accidents on the road today is caused simply due to driver fatigue. Fatigue or otherwise drowsiness, significantly reduces the concentration and vigilance of the driver thereby increasing the risk of inherent human error leading to injuries and fatalities. Hence, our primary motive being - to reduce road accidents using a non-intrusive image processing based alert system. In this regard, we have built a system that detects driver drowsiness by real time tracking and monitoring the pattern of the driver's eyes. The stand alone system consists of 3 interconnected components - a processor, a camera and an alarm. After initial facial detection, the eyes are located, extracted and continuously monitored to check whether they are open or closed on the basis of a pixel-by-pixel method. When the eyes are seen to be closed for a certain amount of time, drowsiness is said to be detected and an alarm is issued accordingly to alert the driver and hence, prevent a casualty. 


\end{abstract}

\section{Introduction}
Recent figures suggest that 109,000 crashes, 87,000 serious injuries and 7,400 deaths in 2017 can be attributed to fatigue related crashes alone. Thus, the implementation and use of measures and systems that can counteract the effect of fatigue and drowsiness is a major obstacle in the domain of vehicular risk management. Yet, we rarely come across such systems in our daily life which suggests that such safety critical systems are understated today. Our aim was to design a fully functional drowsiness detection device which can be ported to any automobile since the need is universal across vehicles. In the design of this system,  crucial requirements that had to be satisfied include real time use with minimal time lag, cost effectiveness of the overall system, non-intrusiveness to the driver and performance in widely varying illumination conditions. Through extensive manual trials and observations under different testing conditions similar to real world driving, credible measures of safety and performance were identified and validated the working of the system and its requirements.   

\subsection{Previous work}
Due to the danger that drowsiness presents on the road, methods have been and are being developed by a multitude of approaches. The whole idea of drowsiness can be detected in a host of ways including sensing physiological characteristics, analysing steering behaviour and monitoring vehicle response. \\
Physiological measurement techniques such as \cite{1} use photoplethysmography (PPG) to analyse signals while \cite{2} extract ECG and EMG signals to detect driver fatigue. Works such as \cite{3} use ECG sensors embedded on the steering wheel wherein the signals are transmitted wirelessly to a server PC for detection. \cite{4} used EEG alpha activity and standard deviation of heart rate (SDNN) were being used to check performance during fatigue tasks.  
Physiological measurement techniques that scan brain activity, heart rate, skin conductance, although being relatively accurate, are often not pragmatic as sensing electrodes would have to be attached directly onto the driver's body and hence, would be quite intrusive and distracting to the driver essentially becoming counteractive. \\
Next, steering behavior based systems such as \cite{5} focus on actions such as slow drifting and corrective counter steering and an ensemble of classifiers to detect fatigue. \cite{6} uses Steering Wheel Angle (SWA) data collected from sensors on the wheel to make decisions. Such systems although being accurate and viable are just not feasible for our proposed system that is aiming towards cost effectiveness. \\
Further, studies such as \cite{7} use driver response feedback monitoring systems. In the long run, such techniques may become tiresome to the driver. \\ 
As a result of the previous work analysed, using non-intrusive computer vision based techniques like monitoring the state of the eyes, extracting facial features and the leaning posture of the driver's head seemed the right direction to progress towards \cite{8,9,10,11,12}. \\
The robustness of the system depends on the quality of image acquisition of the driver's face which is largely dependent on illumination. For this purpose the driver’s face is illuminated using near-infrared (NIR) LEDs on the camera. It minimizes the impact of different ambient light conditions as is typical of real world driving. As the Near-Infrared illumination is not visible to the driver, it is strictly non-intrusive and does not interfere with the driver's ability, which makes our system a lot more suited for real world driving conditions. 

\section{Working}

The working of the drowsiness detection system can be broadly broken down into 3 interdependent stages that hierarchically progress in the given flow : \\
(i)  Eye Centre Tracking \\
(ii) Monitoring state of the eye \\ 
(iii) Drowsiness determination \\

\subsection{Eye Centre Tracking}

Prior to eye center tracking, facial detection needs to be performed first. In our system, this is performed using the reliable Haar Cascades method \cite{13}. Once the face is detected, we instantly extract the RoI and move to the eye region towards the aim of eye center tracking. \\ 
For the purpose of eye detection, the initial method tried was along the lines of template matching \cite{14,15,16}. However, we observed that template matching doesn't allow the driver much flexibility. As is typical in a real world driving scenario, the driver's facial orientation, gaze and posture are constantly in motion. It would then become obtrusive to the driver if asked to be static. The proposed method thus necessitates the need to be able to track the eyes in real time and accommodate for varying gazes and posture. \\
The approach we undertook for constant monitoring of the eye pattern was modelled on the localization of the eye centre (iris) by an image gradient based algorithm. The localization of eye centres was the first major step in our implementation \cite{17,18,19}. This was required to keep track of the eyes throughout and to limit processing of frames to only the eyes of the driver.  

The bi-step process for eye center tracking is :- \\
(i) Mathematically define the center of a circular pattern as the intersection of maximum image gradients. \\
(ii) Use prior knowledge of the eye appearance and location to increase accuracy. \\

Hereafter, we look at the image gradients and apply them to their vector field characteristics. For a probable center of the eye, \textbf{c}, where \textbf{$g_i$} is the gradient vector at position \textbf{$x_i$}, then the normalized displacement vector \textbf{$d_i$} should have the same direction as the gradient vector. The dot product of the normalized displacement vectors and the gradient vectors (for a fixed c) should thus guide us to find the estimated centre of the eye, $c^*$. 

\begin{equation}
 c^* = arg max ( \frac{\sum_{i=1}^{N}(d_i^T.g_i)^2}{N}) \enspace ,
\end{equation}
\begin{equation}
d_i = \frac{(x_i-c)}{||x_i-c||_2}
\quad , \quad \forall i : ||g_i||_2=1 \enspace.
\end{equation}

The displacement vectors d$_{i}$ are scaled to ensure the same weight for all pixel positions. In order to better robustness to variations in illumination and contrast, the gradient vectors should also be scaled to unit length. The maxima of this objective function is highest when the centre estimate is accurate i.e at the iris. However in some cases, there might be local maxima present which leads to inaccurate localization. Since we know that the pupil is darker than the sclera (the white portion surrounding the iris), we can use $c^*$ as a weighted average, where weights are given to each probable centre, c.
\begin{equation}
c^* = arg max ( \frac{\sum_{i=1}^{N} w_c.(d_i^T.g_i)^2}{N}) \enspace,
w_c =I(c_x,c_y) \enspace .
\end{equation}

This weight \textbf{$w_c$} is the grey value at \textbf{($c_x,c_y$)} of the input image I.

\begin{figure}[h]
\centering
\subfloat[Extra light][Extra light]{
\includegraphics[width=3.5cm, height=3cm]{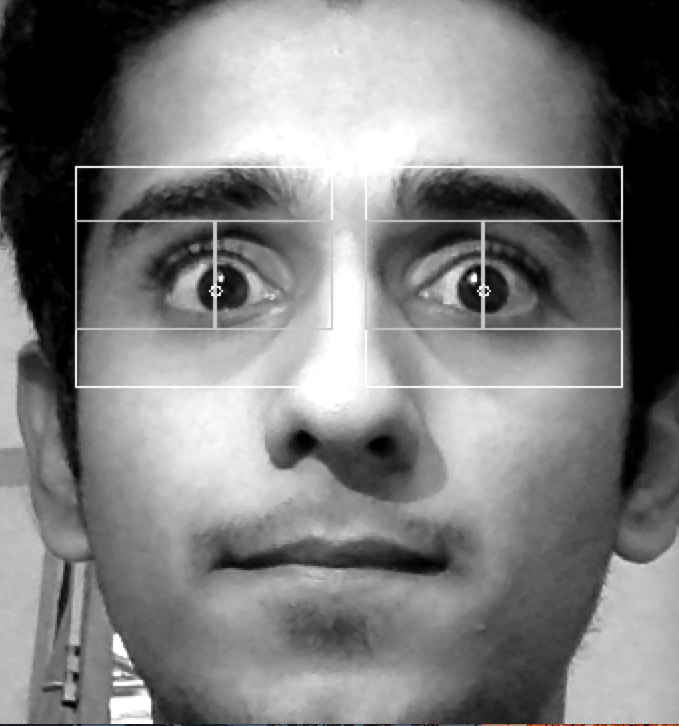}
\label{fig:subfig1}}
\subfloat[Gaze Right Corner][Gaze Right Corner]{
\includegraphics[width=3.5cm, height=3cm]{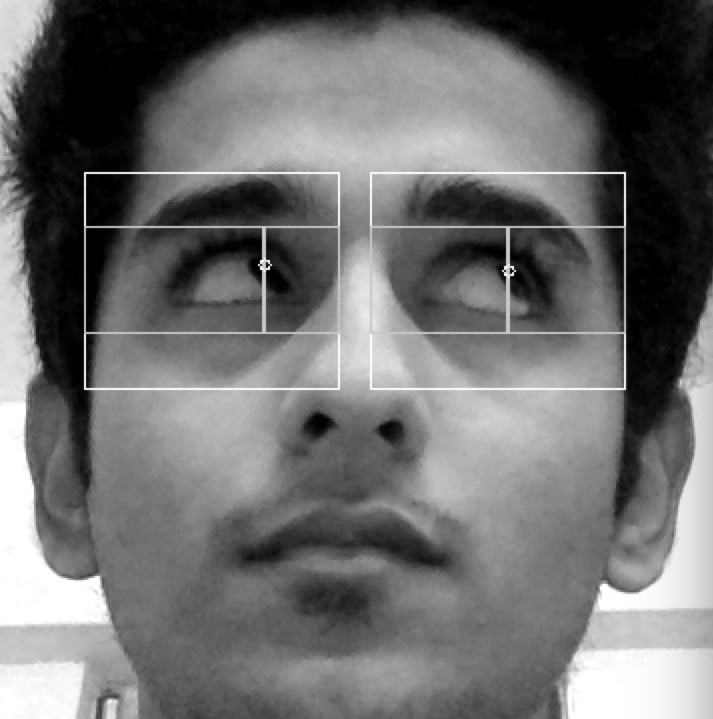}
\label{fig:subfig2}}
\subfloat[Gaze Left Corner][Gaze Left Corner]{
\includegraphics[width=3.5cm, height=3cm]{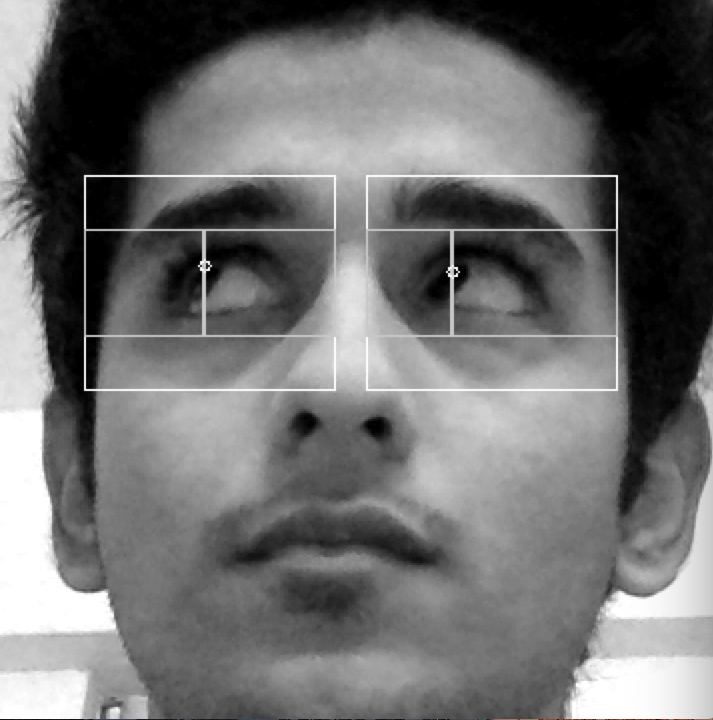}
\label{fig:subfig3}}
\qquad
\subfloat[With Spectacles][With Spectacles]{
\includegraphics[width=3.5cm, height=3cm]{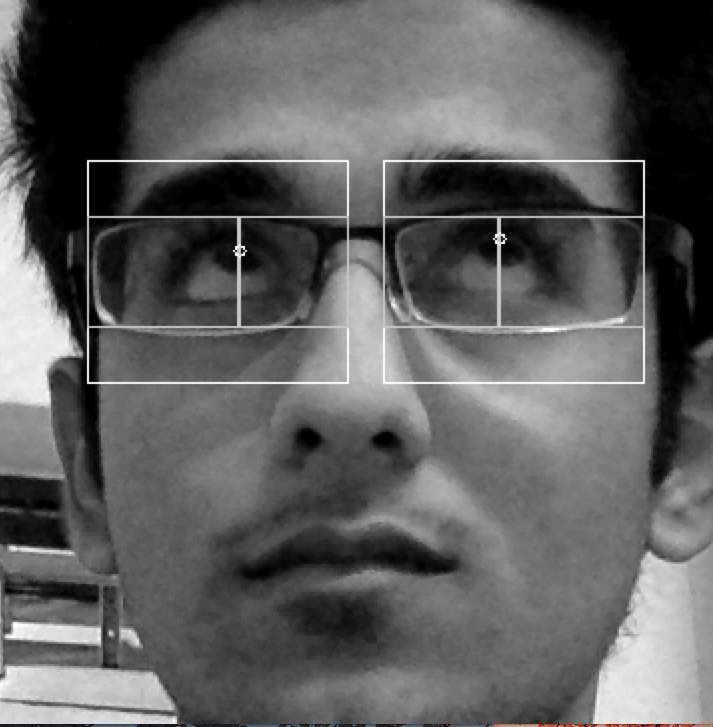}
\label{fig:subfig4}}
\subfloat[Inaccurate due to side view][Inaccurate due to side view]{
\includegraphics[width=3.5cm, height=3cm]{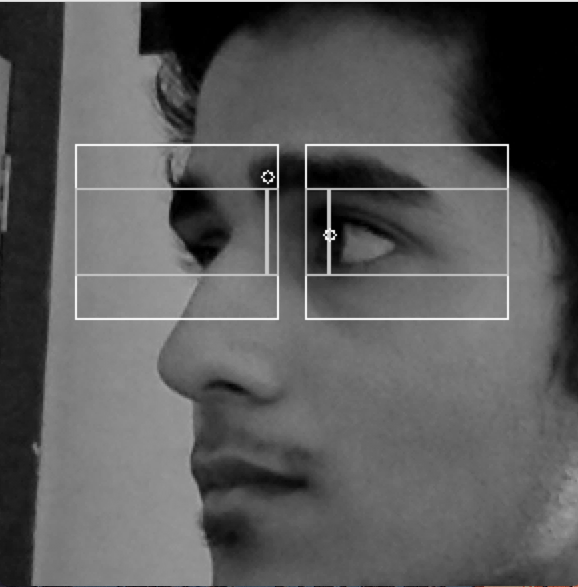}
\label{fig:subfig5}}
\subfloat[Almost Closed eye][Almost Closed eye]{
\includegraphics[width=3.5cm, height=3cm]{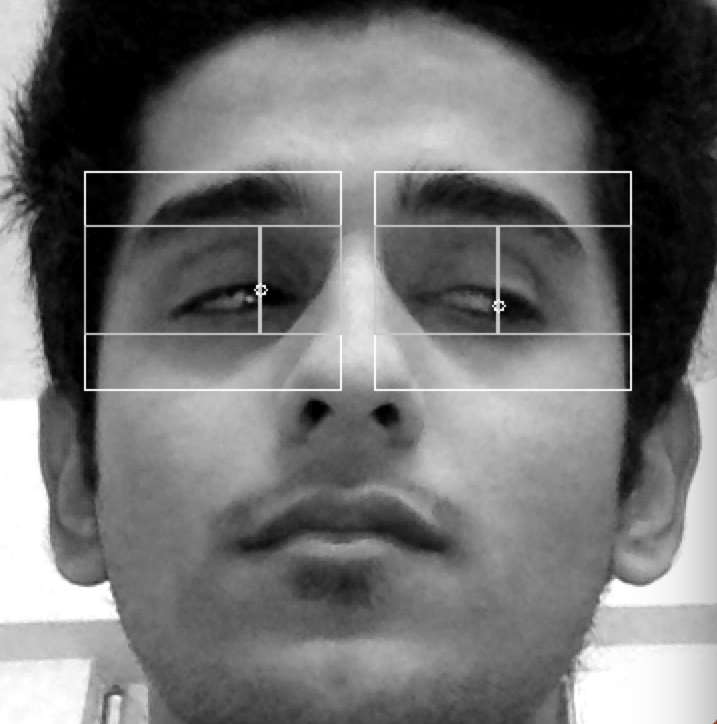}
\label{fig:subfig6}}
\qquad
\caption{Eye centre tracking under varying scenarios.}
\label{fig:globfig1}
\end{figure}

\subsection{Monitoring the state of the eye}

\begin{figure}[h]
\centering
\captionsetup{justification = centering}
\subfloat[Original input image]{
\includegraphics[width=2.5cm, height=2.5cm]{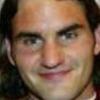}
\label{fig:subfig14}}
\subfloat[Eye centres localized]{
\includegraphics[width=2.5cm, height=2.5cm]{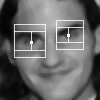}
\label{fig:subfig15}}
\subfloat[Open right eye]{
\includegraphics[width=2.5cm, height=2.5cm]{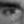}
\label{fig:subfig16}}
\subfloat[Open left eye]{
\includegraphics[width=2.5cm, height=2.5cm]{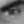}
\label{fig:subfig16}}
\qquad
\subfloat[Original input image]{
\includegraphics[width=2.5cm, height=2.5cm]{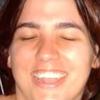}
\label{fig:subfig17}}
\subfloat[Unable to detect eye centers as eyes are closed]{
\includegraphics[width=2.5cm, height=2.5cm]{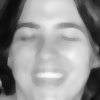}
\label{fig:subfig16}}
\subfloat[Closed right eye]{
\includegraphics[width=2.5cm, height=2.5cm]{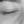}
\label{fig:subfig14}}
\subfloat[Closed left eye]{
\includegraphics[width=2.5cm, height=2.5cm]{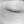}
\label{fig:subfig15}}
\caption{State of the eye detection using the CEW Dataset \cite{20}}
\label{fig:globfig3}
\end{figure}

Once eye centre localization and tracking has been performed successfully as seen in Fig. \ref{fig:globfig1}, we extract real time video windows of the right eye and left eye. This live video is then broken down frame by frame into a series of images. Each image is then binarized on which the analysis occurs. Thereafter, we scan the images of both the eyes obtained from the live window for the percentage of white pixels in each frame. As each image consists of purely black and white, the number of white pixels are determined which gives us the required percentage. After extensive trials in various conditions, of observing the percentage of white in open and closed eyes, it was seen that the percentage of white was greater in closed eyes compared to open eyes. As was observed, during the state of open eyes, the presence of the dark iris led to a lower white percentage. Whereas during the state of closed eyes, light gets reflected onto the closed eyelids, and this surface area region hence, appears as white after the thresholding, leading to the increased amount of white in closed eyes. \\
Of the binary thresholding methods, the objective was to get the maximum difference between the percentage of white in open and closed eyes. After thoroughly examining different techniques (Table. \ref{table:tab1}), Adaptive Gaussian Binary Thresholding (AGBT) was chosen since the percentage difference was highest. Here, different regions of the image have different local thresholds which adds up with the fact that lighting on different regions of the frame are not the same. It was also observed that the relative difference in the percentage of white between open and closed eyes is sustained in this case which can partly be attributed to the Near IR LEDs on the camera. Thus, making AGBT most efficient.
\\ 
To verify how well our detector for open/closed eyes works, we used the demanding CEW dataset \cite{20}, which consists of 2423 images - 1192 closed eye images and 1231 open eye images. The input was a 100x100 pixel coarse face region image as shown in Fig. \ref{fig:globfig3}, similar to the size and quality of the frame recorded from our camera used. The eye center localization is first carried out on the image in greyscale, after which a 25x21 pixel window is extracted of both the eyes for analysis. We thereby obtained an accuracy of 91.6\% on the entire dataset. This seemed like a formidable number to proceed with as the dataset images were characterized by largely varying lighting conditions, blur and shade.

\begin{table}[h]
\begin{center}
\centering
\caption{Binary Thresholding Methods and their relative percentage difference}
\label{table:tab1}
\begin{tabular}{|c|c|}
\hline
Thresholding method & Difference in White Percentage  \\
\hline
Simple Binary & 7\% \\
\hline
Otsu's Thresholding & 8\% \\
\hline
Niblack's Thresholding & 10\% \\ 
\hline
Bernsten's Thresholding & 11.5\% \\
\hline
Mean Gaussian Binary & 13\% \\
\hline
Adaptive Gaussian Binary & 15\% \\
\hline
\end{tabular}
\end{center}
\end{table}

\begin{figure}[h]
\centering
\subfloat[Open Left Eye]{
\includegraphics[width=3cm, height=3cm]{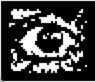}
\label{fig:subfig7}}
\centering
\subfloat[Open Right Eye]{
\includegraphics[width=3cm, height=3cm]{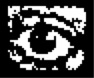}
\label{fig:subfig8}}
\caption{Real time video extracted of the eyes in open state after eye center localization and AGBT}
\label{fig:globfig2}
\end{figure}

\subsection{Drowsiness Determination}
At this point, we have a sequence generating the frame-by-frame percentage of white in the eyes region, the question now arises as to how we would determine when the user is actually classified as drowsy. 
The threshold value that needs to distinguish between a closed eye and an open eye needs to be dynamic and change in accordance with the surroundings to accurately detect the state of the eye. Instances of such situations would be when the vehicle is on an open road and the threshold is high with more light being reflected and then the vehicle enters an underpass tunnel that is dimly lighted with the threshold going down as a consequence. Essentially, there are seldom situations where the external lighting stays constant for a vehicle driver, so the critical value that determines eventual drowsiness needs to consistently change and adapt and therefore be dynamic in nature, to ensure optimal efficiency and accuracy. \\
Using the concept of a circular array with 300 entries, we maintain a moving average of the percentage of white of the last 300 frames at any given point of time. For a 30 FPS camera, this would mean averaging the last 10 seconds at that point. Let's call this value A$_{10}$. We found results to be better on using a weighted moving average to find A$_{10}$ with the higher weights given to the more recent time intervals of two seconds incrementally.   
\begin{equation}
A_{10}(\%) = (1/300)\sum_{i=1}^{5} \sum_{i}[i*0.2*W_{i}] 
\end{equation}
\hspace{30mm} \textit{where $W_i$ is the white percentage in the $i^{th}$ interval of 2s}  
\\
As highlighted in Section 2.2, we observed that the difference in the percentage of white between the open and closed eye in the real time windows is almost constant (15\%) in different conditions after applying AGBT (Fig. \ref{fig:globfig2}). Keeping this in consideration, we could then obtain our Percentage Threshold (PTh) i.e the threshold value to distinguish the state of the eyes, from the value of A$_{10}$ itself. Thus, the PTh would equal (A$_{10}$ + C*15\%) where C $\epsilon$ [0:1].
After performing over 1200 trials with a script that used co-efficient values (C) from 0 to 1 in steps of 0.02, the optimal value that maximised objective accuracy function and minimized time lag was C = 0.72.
\begin{equation}
    PTh(\%) = A_{10} + 0.72*15\%
\end{equation}
Finally, we apply a dynamic double thresholding method to detect driver drowsiness. It needs to be kept in mind that a series of closed eyes leads to drowsiness. Hence, this average of white percentage of the last 300 frames (A$_{10}$) obtained is continuously checked to see if it is greater than PTh and once A$_{10}$ $>=$ PTh, then a counter starts where the counter measures the number of continuous frames where the average has stayed high enough, say K, and once the counter value K crosses a specified number which is our Frame Threshold (FTh), then the person is detected to be drowsy and the alarm rings instantly. Therefore, in this dynamic double thresholding method, our first threshold - PTh and the second threshold - FTh, are both dynamic values which are continually changing with every frame. \\ 

\begin{figure}[h]
\centering
\includegraphics[width=8cm, height=8cm]{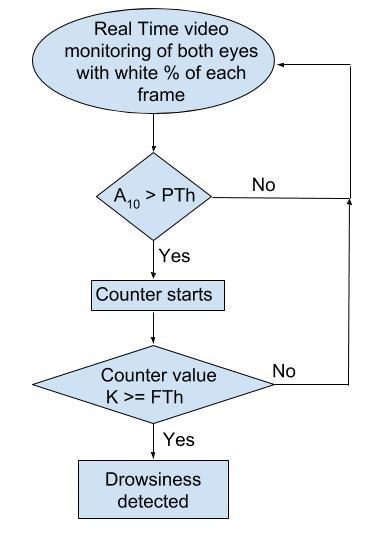}
\label{fig:subfig15}
\qquad
\caption{Drowsiness Determination Algorithm Flowchart}
\label{fig:globfig}
\end{figure}

\section{Features and Conclusions}
The following are some of the crucial features that enable the system to be a pragmatic product and ready for real world usage in vehicles. \\
- The primary objective was developing a cost effective system. The need for drowsiness detection systems and accident avoidance systems in general, is universal across all vehicular types. Fatigue detection systems do exist but only in high end, luxury cars. Hence, a top priority was developing a cost effective and inexpensive system that could be in widespread use, especially in heavy vehicles where the drivers traverse lengthy journeys that are more susceptible to fatigue \cite{21}.   

\begin{table}[h]
\begin{center}
\caption{Hardware specifications of the system}
\label{table:2}
\begin{tabular}{|c|c|c|}
\hline
Component & Specifications & Cost \\
\hline
Raspberry Pi 3 Model B & ARM Cortex-A53 CPU & \$32 \\ 
\hline
Logitech C270 Webcam & Upto 30 FPS at 640x480 & \$15 \\
\hline
RPi 3 compatible Alarm System & 9012 triode driven buzzer & \$1.8 \\
\hline
\end{tabular}
\end{center}
\end{table} 

From Table \ref{table:2}., we see that the cost of the overall system comes up to just under \textbf{\$50}, which is definitely quite affordable considering there are no maintenance costs and the like. Also, the system has no specific pre-requisites per se and can be ported onto any vehicle with a power source, thus satisfying the affordability requirements of the system. \\   
- As the project is primarily concerned with avoiding accident fatalities, having a significant time delay is not an option. A special feature of our system would be its minimal, almost non-existent time lag. The reason for this would be the lack of any training data or machine learning in our system. This would consume massive processing power and as a result, the overall performance of the processor will reduce. There's also no need for any system RAM as nothing is being saved or overwritten in our processing. This further increases the speed and throughput of our system.  \\ After over 1200 trials, the time lag was averaged and seen to be approximately \textbf{0.88 seconds}. With a lag of less than one second, the detection via the alarm system is almost instantaneous. \\
- The compactness and light weight of the system is a huge plus point which makes the product free to move and ready for usage. It can hence be mounted into a vehicle with ease. It can fit into a \textbf{20x15x8 cm\textsuperscript{3} box}. \\ 
- Moreover, the system provides flexibility to the user to really define what drowsiness is to them. The user can easily customize the thresholds for drowsiness to their requirements and set the bar accordingly.

\end{document}